\pdfoutput=1

\documentclass[11pt]{article}

\usepackage[final]{acl}

\usepackage{times}
\usepackage{latexsym}

\usepackage[T1]{fontenc}

\usepackage[utf8]{inputenc}

\usepackage{microtype}

\usepackage{inconsolata}
\usepackage{algorithm}
\usepackage{algorithmic}
\usepackage{amssymb}
\usepackage{dsfont}
\usepackage{amsmath}
\usepackage{graphicx}
\usepackage{multirow}
\usepackage{rotating}
\usepackage{multirow}
\usepackage{booktabs}
\usepackage{array} 
\usepackage{makecell}
\usepackage[normalem]{ulem}
\useunder{\uline}{\ul}{}
\usepackage{url}
\title{IM-BERT: Enhancing Robustness of BERT through the Implicit Euler Method}
\usepackage{authblk}
\usepackage{footmisc}
\usepackage{lipsum} 
\newcommand*{\affmark}[1][*]{\textsuperscript{#1}}
\newcommand*{\email}[1]{\texttt{#1}}
\newcommand\tab[1][1.5cm]{\hspace*{#1}}
\author{\begin{large} \textbf{Mihyeon Kim\affmark[1]}\thanks{\small {Currently at: KT CORPORATION, \email{mihyeon.gim@kt.com}}} \tab  \textbf{Juhyoung Park\affmark[2]}\thanks{\small {Currently at: VAIV COMAPANY, \email{juu9802@vaiv.kr}}} \tab \textbf{Youngbin Kim\affmark[1, 3]} \end{large}\\\\

  \begin{normalsize}\affmark[1]Department of Artificial Intelligence, Chung-Ang University \end{normalsize} \\
  \begin{normalsize}\affmark[2]School of Computer Science and Engineering,
 Chung-Ang University \end{normalsize} \\
  \begin{normalsize}\affmark[3]Graduate School of Advanced Imaging Sciences, Multimedia and Film, Chung-Ang University \end{normalsize} \\
  \begin{normalsize} \email{\{mh10967, wngud51, ybkim85\}.cau.ac.kr} \end{normalsize}\\
  }
\date{}

\begin{document}
\maketitle

\begin{abstract}
Pre-trained Language Models (PLMs) have achieved remarkable performance on diverse NLP tasks through pre-training and fine-tuning. However, fine-tuning the model with a large number of parameters on limited downstream datasets often leads to vulnerability to adversarial attacks, causing overfitting of the model on standard datasets. To address these issues, we propose IM-BERT from the perspective of a dynamic system by conceptualizing a layer of BERT as a solution of Ordinary Differential Equations (ODEs). Under the situation of initial value perturbation, we analyze the numerical stability of two main numerical ODE solvers: \textit{the explicit and implicit Euler approaches.} Based on these analyses, we introduce a numerically robust IM-connection incorporating BERT’s layers. This strategy enhances the robustness of PLMs against adversarial attacks, even in low-resource scenarios, without introducing additional parameters or adversarial training strategies. Experimental results on the adversarial GLUE (AdvGLUE) dataset validate the robustness of IM-BERT under various conditions. Compared to the original BERT, IM-BERT exhibits a performance improvement of approximately 8.3\%p on the AdvGLUE dataset. Furthermore, in low-resource scenarios, IM-BERT outperforms BERT by achieving 5.9\%p higher accuracy.
\end{abstract}

\section{Introduction}

Many Pre-trained Language Models (PLMs) ~\cite{33,37,38} have shown remarkable performances on a wide range of downstream tasks. The high performance of PLMs in domain-specific downstream tasks is achieved through a two-stage training process. In the first stage, PLMs are pre-trained on a high-resource corpus for general representation learning. Subsequently, PLMs are fine-tuned on low-resource data for specific downstream tasks.

On the other hand, this two-stage training process makes PLMs prone to overfitting on the standard dataset, rendering them vulnerable to adversarial attacks through perturbations ~\cite{23,27,29,30,31,32}. These issues arise from aggressively fine-tuning PLMs with a vast number of parameters on low-resource datasets to enhance performance ~\cite{28,36}. Given that PLMs require pre-training on high-resource data to obtain high-quality representations, maintaining the number of parameters is essential. The dilemma suggests addressing the issue during the second step, the fine-tuning stage. 

Numerous studies ~\cite{24,25,28} have been conducted in the field of adversarial training, focusing on the fine-tuning stage. They focus on enhancing robustness using various training strategies, including regularization and optimization techniques in backward propagation with dataset-based hyperparameters. Unlike previous approaches, there is a largely unexplored area that enhances robustness against adversarial attacks in forward propagation on standard training. We propose a robust network architecture for PLMs inherently more resistant to adversarial attacks on standard fine-tuning.

From a dynamic system perspective, a plethora of studies~\cite{12,13,19} have explored the relationship between neural networks and Ordinary Differential Equations (ODEs), interpreting the residual connection within neural networks as an Euler discretization of ODEs. This insight facilitates the integration of various numerical ODE solvers into neural networks. The numerical solvers can be categorized into two approaches: \textit{explicit and implicit approaches.} Among the two approaches, the implicit method itself exhibits high stability against perturbations, such as adversarial attacks~\cite{8}. By adopting the implicit method, several studies~\cite{1,2,4,11} have demonstrated stable performance against both white-box and black-box adversarial attack methods in vision tasks.

Inspired by these previous works, we analyze the theoretical stability of two kinds of numerical ODEs solvers. We consider adversarial attacks as a kind of perturbation on inputs and compare the numerical stability of the two approaches. Based on theoretical analysis, we propose the IM-connection, utilizing the implicit Euler method, to construct a robust architecture against perturbations. By incorporating IM-connection within BERT, we develop IM-BERT to mitigate vulnerabilities to adversarial attacks. This analysis enables us to establish a theoretical basis to construct robust neural architectures without necessitating specialized training strategies or an increased number of parameters.

To validate that our method effectively defends against a wide range of adversarial attacks, we conduct experiments using the Adversarial GLUE (AdvGLUE) dataset~\cite{6}. Our model outperforms the baseline by up to 4.5\%p on average across all tasks in AdvGLUE. Moreover, under stringent conditions with sampled instances, IM-BERT exhibits higher accuracy than BERT, with an improvement of up to 5.9\%p on average. To mitigate the time-cost problem, we demonstrate that an IM-BERT model, applying the IM-connection between several layers, is comparable to one with full implicit connections among layers.

Our main contributions are:

\begin{itemize}
    \item We analyze BERT as an ODE solver and introduce IM-BERT with the IM-connection in an implicit approach. To the best of our knowledge, this is the first work to consider PLMs and enhance their robustness against adversarial attacks from a dynamic system perspective.\vspace{-1mm}
    \item We demonstrate that IM-BERT reinforces resistance to adversarial attacks that perturb at diverse levels. This approach only requires simple modification in standard training.
    \item We validate its robustness even in low-resource scenarios, suggesting that IM-BERT can maintain stability even when training on scarce datasets.
\end{itemize}

\section{Related Work}
\subsection{Robust Learning against Adversarial Attack}
Numerous studies~\cite{28, 40, 41, 24} have devised training strategies aimed at defending against adversarial attacks. One prominent approach is adversarial training, which seeks to produce correct predictions by incorporating adversarial perturbations into the learning process using regularization and optimization techniques.

In terms of regularization, SMART~\cite{28} proposes robust fine-tuning using a regularization module and optimization derived from the proximal point method. This approach highlights that overfitting is likely to occur when PLMs are fine-tuned on limited downstream data. Similarly, R3F and R4F~\cite{24} fine-tune PLMs rooted in an approximation to trust region, preventing degradation of generalizable representations of PLMs.

To optimize adversarial samples, FreeLB~\cite{40} employs a gradient method that perturbs word embeddings to minimize adversarial risks during adversarial training. Like FreeLB, CreAT~\cite{25} effectively guides the optimization of adversarial samples to deceive the entire model more efficiently.

Training on limited adversarial samples not only makes defending against various attacks challenging but also necessitates the generation of costly adversarial perturbations. Moreover, they do not directly address the intrinsic robustness of the architecture itself. PlugAT~\cite{44} has explored the design of inherently robust network architectures by introducing specific modules that enable robust predictions with only a modest increase in the number of trainable parameters. Despite these efforts, the realm of adversarial defense through robust architecture design in forward propagation remains largely unexplored, to our knowledge. 

Motivated by the need for a robust architecture that defends against adversarial attacks, we aim to design a resilient architecture for standard fine-tuning. This is achieved even in challenging conditions, by fine-tuning models solely on clean data, focusing on the inherent stability of the architecture. 

\subsection{ODE Neural Network}
The relationship between neural networks and Ordinary Differential Equations (ODEs) has been elucidated from various perspectives~\cite{13, 5, 3}. These studies demonstrate that the architecture of ResNet~\cite{50} and RNN~\cite{51} can be interpreted as an explicit Euler discretization: the first-order explicit Euler method~\cite{12,13,19}. Based on this interpretation, robust and efficient models have been proposed, utilizing ODE numerical solvers: explicit and implicit approaches.

With explicit methods, the majority of research has focused on achieving stable architectures by emphasizing higher-order precision or step size.  The first-order accuracy of residual connection leads the neural networks to accumulate larger truncation errors~\cite{17, 46, 18, 22}. One stable architecture~\cite{46} shows stable forward and backward propagations by employing the Leapfrog ODE solver and the Verlet method. Similarly, in NLP, the ODE Transformer~\cite{47} applied the Runge-Kutta method, showcasing its stability by obtaining more precise hidden states and preventing accumulated errors. On the other side, scaling the output of layers by the step size in the residual connection contributes to generating more stable predictions against Gaussian noise~\cite{16}. By setting the step size to less than 1 or a learnable parameter, the models satisfy the stability region of the explicit method~\cite{14}.

Theoretically, the explicit method has an inherent limitation when the input data is perturbed. The explicit approach may not converge to a solution depending on the step size~\cite{9}. This implies vulnerability to adversarial attacks that deceive the model by perturbing clean data. In contrast, the implicit method exhibits superior numerical stability against adversarial attacks. More robust and stable architectures have been explored by applying the implicit Euler method to residual connections. Previous works have implemented the implicit method in vision by adapting Newton's iteration or gradient descent methods in practice ~\cite{11, 1, 2}. These studies have demonstrated robustness against noise and adversarial attacks. 
However, given the non-differentiable and discrete characteristics of text, along with its diverse nature or adversarial attack, a more versatile approach is required to defend against complex attacks~\cite{23}. Consequently, in extending the research connecting ODEs and neural networks in NLP, we apply the implicit method to BERT, inspired by IE-skips~\cite{1}.

\section{Method}
In this section, we first describe the network as an ODE solver for the initial value problem and compare the numerical stability of the explicit and implicit Euler methods. While the explicit method offers simplicity and speed advantages, it diverges the hidden states when an input is perturbed. Conversely, the implicit method shows absolute stability compared to the explicit method. With this insight, we opt for the implicit method as an ODE solver. Leveraging the gradient descent method to implement the implicit method, we introduce IM-connection and IM-BERT, a more stable approach for PLMs.
\subsection{Neural Network as an ODE Solver}
Given a dataset with $\left\{X,Y\right\}=\left\{(x,y)|x \in \mathbb{R}^{n}, y \in C \right\}$, a neural network maps $x$ to $y$ as a mapping function $\Phi: X \to Y$. The network consists of $n$ layers, which are nonlinear transformations $\phi_t : H_t \times \Theta_t \to H_{t+1}$ parameterized by hidden states $h_t \in H$ and parameters $\theta_t \in \Theta_t$ in $t$-th layer. In general terms, the hidden state $h_t$ is derived from the $t$-th layer with the input $h_{t-1}$ and $\theta_{t}$, and is governed by the following equation:
\begin{flalign*} 
h_t=\phi_{t}(h_{t-1},\theta_t)
&&(h_0=x, t=1,…,N) 
\tag{1}\label{eq:1}
\end{flalign*} 
Theoretically, as the neural network becomes very deep (i.e., as $t$ approaches $\infty$), we can formulate the network for continuous $t$ as:
 \begin{flalign*}
    h(t)=\phi(x, \theta(t)) \qquad t>0 \tag{2}\label{eq:2}
 \end{flalign*}  
From the perspective of ODEs, we can establish a connection between neural networks and ODEs, where a change of hidden states in the network for $t$ is related to the layer of the network. In the continuous flow of $t$, the hidden states $h$ vary through the layer. As $t$ flows, the process of changing hidden states $h(t)$ through layers can be formulated into the following ODE:
 \begin{flalign*}
     \frac{dh}{dt}= \phi(h(t), \theta(t)) \tag{3}\label{eq:3}
 \end{flalign*} 
Given $\phi(x, \theta(t))$, the network seeks all possible hidden states $h(t)$ that enable it to predict the final hidden states $y$ while satisfying Eq. 3 via input $x$, the initial value of the hidden states. Therefore, a neural network with $n$ layers can be interpreted as an ODE solver to find the solution to the initial value problem:
 \begin{flalign*}
     \frac{dh}{dt} = \phi(h(t), \theta(t)) \quad h(0)=x \ ,\ t>0 \tag{4}\label{eq:4}
 \end{flalign*}  
\subsection{Residual Connection as Euler Method}
Directly solving the initial value problem of Eq. 4 is generally infeasible. Therefore, it is necessary to use a numerical procedure to approximate a solution. The right side of Eq. 4 is approximated by forward and backward difference methods for the range from $t$ to $t+\gamma$ and from $t-\gamma$ to $t$, with a step size $\gamma$. Applying these methods, both explicit and implicit Euler methods find the hidden states $h_t$ as a solution using layers $\phi_t$ and the output $h_{t-1}$ of the previous layer, sequentially.\\
\\
\textbf{Explicit Method} 
Obtaining $h_t$ is a simple process of updating the previous hidden states $h_{t-1}$ by adding the product of the step size $\gamma$ and the transformation $\phi_{t}(h_{t-1}, \theta_t)$ through the $t$-th layer. The computation $h_t$ is written as:
\begin{flalign*}
    h_t=h_{t-1}+\gamma\ \phi_t(h_{t-1},\theta_t )\\ 
    h_0 = x,\ t=1,...,N \tag{5} \label{eq:5}
\end{flalign*}
When the step size $\gamma$ equals 1, the explicit Euler method serves as the residual connection in the neural network.\\
\textbf{Implicit Method}
To compute $h_t$ in the implicit method, it is essential to solve the following nonlinear equation:
\begin{flalign*}
    h_t=h_{t-1}+\gamma\ \phi_{t}(h_t, \theta_t) \\
    h_0 = x,\ t=1,...,N \tag{6} \label{eq:6}
\end{flalign*}
assuming that $\phi_t(h_t, \theta_t)$ is Lipschitz continuous, this equation has a unique solution if the step size $\gamma$ is sufficiently small. 

Both the explicit and implicit systems generate hidden states inductively, starting with the initial input $x$. Thus, it is critical to understand whether the two systems can stably approximate the hidden states when perturbations are added to $x$. 

\subsection{Stability of Implicit Connection in Perturbation}
We first define stability and stable regions when the initial value $x$ is perturbed. We then analyze the stability of each method using the model equation, which is generally used for stability evaluation: 
\begin{flalign*}
    \frac{dt}{dh}=\lambda h(t)+\psi(t,x)\quad x=h(0)\ ,\ \lambda <0  \tag{7}\label{eq:7}
\end{flalign*}
\textbf{Definition 1. (Absolute Stability)}
The system is absolutely stable for its initial value problem with input $x$ if there exists a step size $\gamma$ such that the error between hidden states from $x+\eta$ and $x$ converges to zero:
\begin{flalign*}
    \lim_{n \to \infty}(\phi_n(x+\eta, \theta_n)-\phi_n(x, \theta_n)) = 0  \tag{8}\label{eq:8}
\end{flalign*} 
\textbf{Definition 2. (Region of Absolute Stability)}
The set of all $\lambda \gamma$, in which $\lambda, \gamma$ satisfy absolute stability, is called the region of absolute stability. \\

As shown in the two definitions above, when the initial value is perturbed, the region of absolute stability becomes constrained,  as it must satisfy absolute stability. We analyze the region of absolute stability in the implicit and explicit methods in the following propositions:\\

\noindent\textbf{Proposition 1. (Stability of Explicit Method)}
For an explicit Euler method, the model equation is absolutely stable if and only if the region of absolute stability is $|1+\gamma\lambda| <1$. \\
\begin{figure*}[t]
\centering
\includegraphics[width=0.9\textwidth]{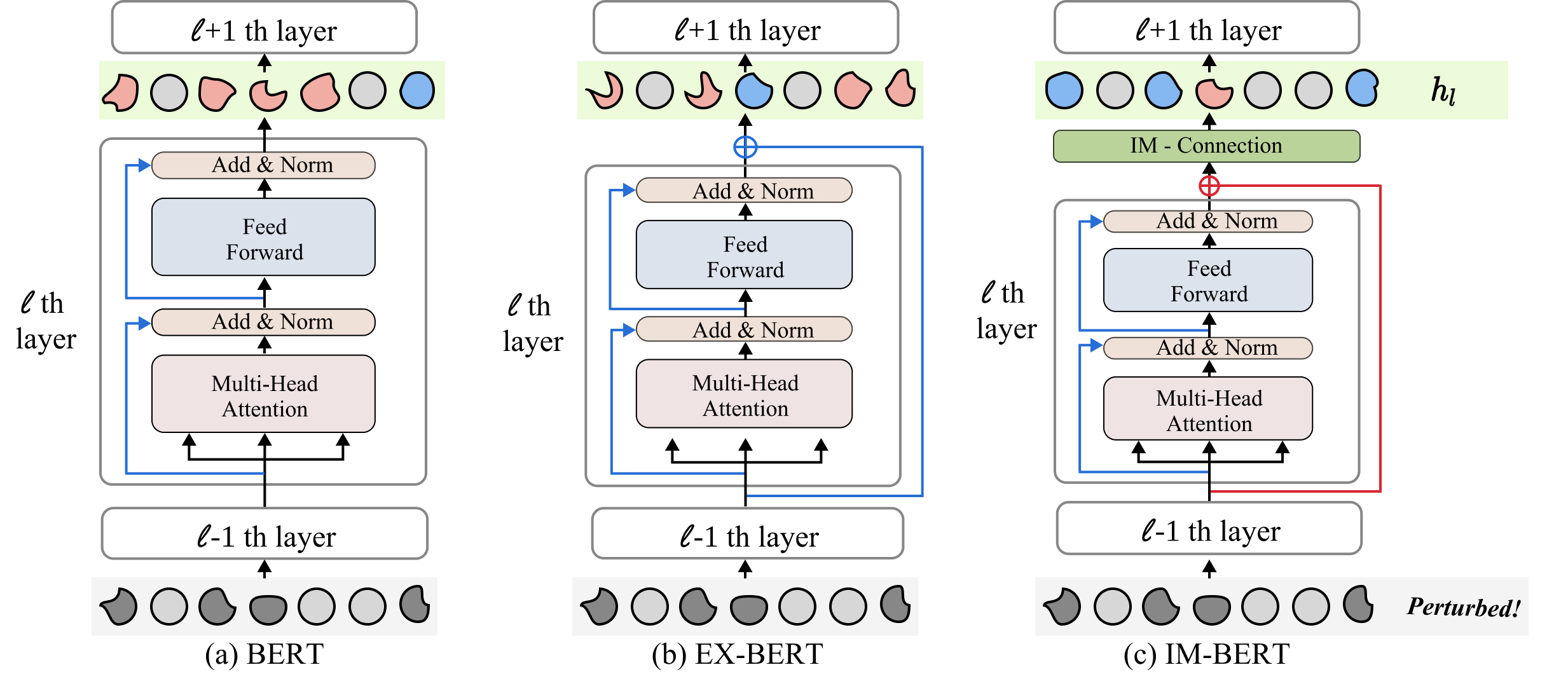} 
\caption{\textbf{The $l$-th layer in each architecture} The perturbed inputs pass through the layer to estimate the hidden states. Blue represents the corrected hidden state, while red indicates the opposite. Blue connections represent explicit residual connections, while red connections represent IM-connections. (a) \textbf{BERT}  (b) \textbf{EX-BERT} In the layer, the layers are connected with the explicit method. (c) \textbf{IM-BERT} The hidden states taken from the $l$-th layer are updated in IM-connection.}
\label{fig1}
\end{figure*}
\sloppy\\
\indent\textbf{Proof 1.} Let the solutions of the model equation be $h$ and $h_{\eta}$ when the input is $x$ and $x+\eta$ perturbed by $\eta$. Subtracting the model equation by the equation whose initial value is perturbed, the error of two solutions can be represented as the initial value problem $h_{\eta}-h = \frac{d}{dt}(h_{\eta}-h)$ with $h_{\eta}(0)-h(0)=\eta$. Applying the explicit method, the error becomes $(h_{\eta}-h)_n=(1+\gamma \lambda)^{n} \eta$ when $t=n$ by inductive argument. This error goes to 0 iff $|1+\gamma \lambda|<1$.  $\square$\\
\sloppy\\
\noindent\textbf{Proposition 2. (Stability of Implicit Method)} For an implicit Euler method, the model equation is absolutely stable, regardless of the step size $\gamma$.\\
\sloppy\\
\indent\textbf{Proof 2.} Let the solutions of the model equation be $h$ and $h_{\eta}$ when the input is $x$ and $x+\eta$ perturbed by $\eta$. Subtracting the model equation by the equation whose initial value is perturbed, the error of two solutions can be represented as the initial value problem $h_{\eta}-h =\frac{d}{dt}(h_{\eta}-h)$ with $h_{\eta}(0)-h(0)=\eta$. Applying the implicit method, the error becomes $(h_{\eta}-h)_n=\frac{1}{(1-\gamma \lambda)^{n}} \eta$ when $t=n$. The error goes to 0, regardless of the value $\lambda\gamma$. $\square$  

Therefore, for the initial value $x$ perturbed by $\eta$, the explicit method can diverge in the process of approximating the hidden states $h_t$ in the $t$-th layer, depending on the step size $\gamma$ and the stability region. In contrast, the system to which the implicit method is applied exhibits absolute stability with respect to the perturbed $x$. Proposition 2 indicates that the hidden states are approximated stably, regardless of step size $\eta$. From the perspective of adversarial attacks, which degrade the performance of the model by introducing data perturbation $\eta$, the system with implicit methods can reliably obtain hidden states. These propositions suggest that this network is more robust against adversarial attacks than the explicit one.
\subsection{IM-BERT}
We design the network to guarantee more robust hidden states, acting as an ODE solver, especially in situations where it must counter adversarial attacks. As observed with the residual connection, the explicit method has the advantage of a quick and simple forward process. However, according to the above analysis, it may be challenging for the system to remain within the stability region practically, making it potentially vulnerable to adversarial attacks that perturb the data.
\begin{flalign*}
    h_{t}^*:= arg \min_{x}||h_{t}-h_{t-1}-\gamma \phi_t(h_t , \theta_t )||^2 \tag{9} \label{9}
\end{flalign*} 
\begin{algorithm}[h]
\caption{IM connection within IM-BERT layers}
\label{alg:algorithm}
\textbf{Input}: Hidden states $h_{t-1}$from previous $(t-1)$-th layer and $t\text{-th}$ layer $\phi_{t} (\cdot)$\\
\textbf{Parameter}: Iteration number \textit{T} and step size $\gamma$. (We set a step size $\gamma$ to 0.1 in our experiments.)\\
\textbf{Output}: Hidden states $h_{t}$

\begin{algorithmic}[1] 
    \STATE Initialize $h_{t}^0=h_{t-1}+\phi_t(h_{t-1}, \theta_{t})$
    \FOR{$i=0,...,T-1$}
    \STATE $loss_{i}=||h_{t}^{i}-h_{t-1}-\phi_t(h_{t}^{i})||^2$
    \STATE Compute gradient $\bigtriangledown_{h_{t}^{i}}loss_i$.
    \STATE $h_t^i=h_t^i-\gamma \bigtriangledown_{h_{t}^{i}}loss_i$
    \ENDFOR
    \STATE \textbf{return} The hidden states $h_t$ of $t$-th layer.
\end{algorithmic}
\end{algorithm}
To address this issue, we modify the residual connection by applying an implicit method to ensure the network maintains a stable process. Unlike the residual connection, the implicit method requires an additional process to find the solution to Eq. 6 to obtain the hidden states. Typically, this solution is found using Newton's iteration~\cite{11, 45}. However, although this approach may be stable for precise input $x$, it no longer guarantees absolute stability as it threatens stability when input is perturbed~\cite{9}.

\begin{table*}[t!]
\centering
\setlength{\tabcolsep}{6pt}
\renewcommand{\arraystretch}{1.0}
\resizebox{\textwidth}{!}
{%
{\tiny
\begin{tabular}{ll|cccccc|c}
\toprule[0.6pt]\hline
 &  & SST-2 & QQP & QNLI & MNLI-m & MNLI-mm & RTE & \textit{Avg.} \\ \hline
\multicolumn{1}{l|}{\multirow{4}{*}{Adversarial Training}} & FreeLB & 31.6 & 51 & \textbf{45.4} & 33.5 & 21.9 & 42.0 & 37.6 \\
\multicolumn{1}{l|}{} & CreAT & \textit{35.3} & \textbf{51.5} & 34.9 & 36.0 & 22.0 & \textbf{45.2} & \textit{39.1} \\ 
\multicolumn{1}{l|}{} & R3F & 20 & 38.2 & 34.9
& 30.8 & 33.1 & 32.2 & 31.5 \\ 
\multicolumn{1}{l|}{} & SMART & 20.6 & 39.1 & 32.3 & 28.6 & 32.1 & 28.6 & 30.2 \\ \cline{1-9} 
\multicolumn{1}{l|}{\multirow{3}{*}{Standard Training}} & BERT & 31.1 & 47.4 & 36.5 & 33.1 & 27.8 & 34.6 & 35.1 \\
\multicolumn{1}{l|}{} & EX-BERT & 33.8 & 42.3 & 40.5 & \textit{36.3} & \textit{33.3} & 28.4 & 35.8 \\ \cline{2-9} 
\multicolumn{1}{l|}{} & IM-BERT & \textbf{39.2} & \textit{48.8} & \textit{41.2} & \textbf{38.1} &\textbf{35.2} & \textit{44.4} & \textbf{41.2} \\ \hline

\end{tabular}%
}
}
\\

\resizebox{\textwidth}{!}
{%
{\small
\setlength{\tabcolsep}{7pt}
\renewcommand{\arraystretch}{1.1}
\begin{tabular}{ll|cccccc|c}
\hline
 &  & SST-2 & QQP & QNLI & MNLI-m & MNLI-mm & RTE & \textit{Avg.} \\ \hline

\multicolumn{1}{l|}{\multirow{4}{*}{Adversarial Training}} & SMART$_{\textit{large}}$ & 25.21 & 36.49 & 34.61 & 26.89 & 23.32 & 38.16 & 30.29 \\ \cline{2-9}
\multicolumn{1}{l|}{} & SMART$_{\textit{RoBERTa}}$ & 50.29 & 64.22 & 52.17 & \textit{45.56} & 36.07 & \textbf{70.39} & \textbf{53.71} \\ 
\multicolumn{1}{l|}{} & infoBERT$_{\textit{RoBERTa}}$ & 47.61 & 49.29 & 54.86 & \textbf{50.39} & \textbf{41.26} & 39.47 & 46.04 \\ 
\multicolumn{1}{l|}{} & $FreeLB_{\textit{RoBERTa}}$ & \textbf{61.69} & 42.18 & \textbf{62.29} & 31.59 & 27.60 & \textit{62.17} & 50.47 \\ \cline{1-9}
\multicolumn{1}{l|}{\multirow{4}{*}{Standard Training}} & BERT$_{\textit{large}}$ & 33.03 & 37.91 & 39.77 & 28.72 & 27.05 & 40.46 & 34.49 \\
\multicolumn{1}{l|}{} & BERT & 21.48 & 41.23 & 33.47 & 28.59 & 28.69 & 40.79 & 32.38 \\ \cline{2-9} 
\multicolumn{1}{l|}{} & IM-BERT & 44.34 & 36.02 & 48.86& 32.25 & \textit{38.89} & 43.92 & 40.71 \\ 
\multicolumn{1}{l|}{} & IM-RoBERTa & \textit{52.04} & \textbf{69.91} & \textit{56.68}& 33.68 & 37.65 & 60.49 & \textit{51.74} \\ \hline \bottomrule[1pt]
\end{tabular}%
}
}

\caption{\textbf{Results on AdvGLUE} (Upper) For dev sets, We report the accuracy, comparing adversarial training and standard training. Adversarial training results are sourced from the CreAT paper. (Bottom) For test sets, We report the accuracy of adversarial training and standard training, with adversarial training results from the AdvGLUE paper.}
\label{Table1}
\end{table*}
Therefore, inspired by previous works~\cite{1,2}, we adopt a gradient descent algorithm to approximate the solution of Eq. 6, i.e., the hidden states. The aim of this algorithm is to find the optimal hidden states that minimize the transformation represented by Eq. 6.


When initializing $h_t$ in Algorithm 1, the initial value $h_{t}^{0}$ is estimated as an explicit method in terms of ensuring the advantages of forward and backpropagation, like in a previous work~\cite{1}. Afterward, IM-connection finds the optimal hidden states $h_{t}^{*}$ by calculating the gradient of Eq. 9 in a series of iterations \textit{T}. 

Utilizing the gradient descent method to the implicit approach, We compare the three architectures. Firstly, as shown in Figure 1(a), BERT does not have a residual connection between each layer. BERT goes through the process of monotonically transferring hidden states $h_t$ between layers.

The network can be interpreted as an ODE solver and applied as a layer. Inspired by the analysis, as shown in Figure 1(b) and (c), we propose EX-BERT and IM-BERT that transform the connection between layers into explicit and implicit Euler methods, respectively. We implement EX-BERT by adding residual connections between layers in BERT. It seems similar to IB-BERT~\cite{7}. But for a fair comparison with BERT and IM-BERT, we apply only the residual connection among the layers except for layer normalization ~\cite{48}. 

We construct IM-BERT by incorporating BERT and the IM-connection using the gradient descent algorithm. In particular, we choose not to transform all residual connections linking feed-forward and multi-head attention. This decision stems from our consideration of distinct properties between different modules and the trade-off between robustness and time cost. Even if robustness is enhanced through fine-tuning with IM-connection, applying IM-connection to all residual connections causes excessive time costs. Above all, it is experimentally proven that only the IM-connection between layers has significant robustness, even in harsh situations such as a low-resource scenario.

Also, the substitution of IM-connection for the monotone connection implies its applicability to PLMs employing monotone connection interlayers. This integrated operation within layers effectively boosts the robustness of PLMs without increasing the number of parameters.
\section{Experiment}

In this section, we conduct experiments in various adversarial attacks and resource situations to demonstrate the inherent robustness of IM-BERT.
\begin{table*}[h!]
\centering
\resizebox{\textwidth}{!}
{%
{\tiny
\renewcommand{\arraystretch}{0.92}
\setlength{\tabcolsep}{9pt}
\begin{tabular}{l|cccccc|c}
\toprule[0.6pt]
\hline
N=1000        & SST-2 & QQP  & QNLI        & MNLI-m & MNLI-mm & RTE  & \textit{Avg.}  \\ \hline
R3F   & 30.4 & 37.2 & 45.3     & 31.5   & 31.5    & 37.0 & 35.5 \\
SMART   & 31.1 & \textbf{46.2} & 45.9        & 17.4   & 30.9    & 34.2 & 34.3 \\
FreeLB   &33.8& 41.0 & 41.9        & 24.8   & 29.0    & 40.7 & 35.2 \\\hline
BERT   & 32.21 & 33.8 & 43.5        & 32.5   & 31.3    & \textbf{42.4} & 36.0 \\
IM-BERT$_{\textit{T=5}}$  & 39    & 41.5 & 47.1        & \textbf{38.3 }  & 30.9    & 31.3 & 38.0 \\
IM-BERT$_{\textit{T=10}}$ & \textit{39.4}  & 38.0   & \textit{52.9}        & \textit{36.9}   & \textbf{36.4}    & 35.0   & \textit{39.8} \\
IM-BERT$_{\textit{T=15}}$ & \textbf{44.1}  & \textit{45.3} & \textbf{55} & 36.1   & \textit{34.2}    & 36.2 & \textbf{41.8} \\ \hline

\end{tabular}
}
}
\resizebox{\textwidth}{!}
{%
{\tiny
\renewcommand{\arraystretch}{0.95}
\setlength{\tabcolsep}{9pt}
\begin{tabular}{l|cccccc|c}
\toprule[0.4pt]
\hline
N=500       & SST-2 & QQP  & QNLI        & MNLI-m & MNLI-mm & RTE  & \textit{Avg.}  \\ \hline
R3F   & 29.7 & 35.9 & 37.8        & 25.6   & 32.1    & \textbf{49.4} & 35.1 \\
SMART   & 35.1 & 34.6 & 41.9        & 24.0   & 33.3    & 40.7 & 35.0 \\
FreeLB   & 31.8 & 35.9 & \textit{48.0}        & 20.7   & 30.3    & \textit{48.2} & 35.8 \\\hline
BERT & 34 & 32.5 & 43.2 & 25.1 & \textbf{37} & 47.3 & 36.5 \\
IM-BERT$_{\textit{T=5}}$& 41 & 41.9 & \textit{48.0} & 35.8 & 34.2 & 32.9 & 38.9 \\
IM-BERT$_{\textit{T=10}}$ & \textit{41.9} & \textbf{48.3} & \textbf{49.1} & \textbf{39.7} & \textit{36.4} & 39.1 & \textbf{42.4} \\ 
IM-BERT$_{\textit{T=15}}$ & \textbf{45.9} & \textit{44.4} & 47.3 & \textit{39.1} & 32.5 & 36.2 & \textit{40.9} \\ \hline
\bottomrule[0.6pt]
\end{tabular}%
}
}
\caption{\textbf{Results on AdvGLUE under low-resource scenarios.} We report accuracy to evaluate the performance of IM-BERT. (Upper) Results for 1,000 instances. (Bottom) Results for 500 instances. The best performance is highlighted in bold.}
\label{table2}
\end{table*}
\subsection{Robustness against Various Adversarial Attacks}
\textbf{Setup} To evaluate the robustness, we utilize Adversarial GLUE (AdvGlue)~\cite{6}, a robustness benchmark. AdvGLUE comprises various adversarial attacks for some tasks of GLUE~\cite{49}. In public, the dataset includes a substantial number of adversarial samples encompassing word-level, sentence-level perturbations, and human-crafted examples. In our experiments, we fine-tune IM-BERT for standard GLUE and evaluate the robustness for AdvGLUE. We employ our baseline as bert-base-uncased. Both IM-BERT and EX-BERT are warm-started on the baseline. To make a fair comparison, we configure the same hyperparameters as BERT. During fine-tuning the model with the Adam~\cite{43}, we set the learning rate to 2e-5 for all tasks. Additionally, we use a linear warm-up scheduler and set the batch size to 16. The standard GLUE data is split in an 8:2 ratio for each task. The model with the lowest validation loss is selected and evaluated by the GLUE metric in the hugging face. In our model, the hyperparameter iteration \textit{T} and step size $\gamma$ are 5 and 0.1, respectively, in the main experiment.\\
\textbf{Result} Our analysis of IM-BERT's robustness is further validated by its stable performance on AdvGLUE. When compared to EX-BERT and BERT, IM-BERT consistently outperforms, with results on validation sets showing suppression of BERT by up to 9.8\%p and EX-BERT by up to 16\%p in RTE. Averaged across all AdvGLUE tasks, IM-BERT exhibits a 5.4\%p higher accuracy than EX-BERT. Even on test sets, IM-BERT demonstrates its stability, yielding an 8.3\%p higher accuracy than BERT.
Comparing IM-BERT with standard training against adversarial training, we utilize results from AdvGLUE and the CreAT paper. On test sets, IM-BERT in standard fine-tuning outperforms baseline by up to 8.33\%p on average. Additionally, comparisons with other models show that IM-BERT performs similarly or superiorly to BERT$_{\textit{large}}$ and adversarial training methods on BERT$_{\textit{large}}$, achieving up to 9.33\%p higher accuracy over SMART$_{\textit{large}}$~\cite{28} on average. These results affirm IM-BERT's robustness against perturbations.
Table 1 shows that IM-RoBERTa performs well across multiple tasks, notably achieving high scores in SST2, QQP, QNLI, and RTE. Its average performance is higher than FreeLB by up to 1.27\%p and significantly higher than InfoBERT by up to 5.7\%p, indicating it generally outperforms other methods. Specifically, IM-RoBERTa achieves the highest score in QQP and second-place solid finishes in other tasks, demonstrating robust overall performance. RTE and MNLI show relatively lower improvements because IM-RoBERTa focuses on robustness against adversarial perturbations rather than reasoning abilities. Consequently, IM-RoBERTa excels in tasks like QQP while maintaining competitive performance elsewhere. Compared to traditional adversarial training methods, this approach reduces the need for extensive hyperparameter tuning by relying on standard fine-tuning. Thus, the IM-connection enhances inherent robustness with simple architectural modifications.\vspace{-5mm}
\begin{table*}[t]
\setlength{\tabcolsep}{10pt}
\renewcommand{\arraystretch}{1.1}
\centering
\resizebox{\textwidth}{!}{%
{\tiny
\begin{tabular}{c|ccc|c|c}
\toprule[0.7pt]
\hline
& Word & Sentence & Human & All & \makecell[c]{FLOPs / Params} \\ \hline
BERT         & 30.1($\pm$8.8)       & 12.2($\pm$7.1)           & 24.4($\pm$3.5)          & 29.1($\pm$2.3) & \makecell[c]{22.35 / 109.48M} \\ \cline{1-6}
Layer (1-3)   & \textbf{46.2($\pm$8.0)}       & 29.4($\pm$3.1)           & 28.4($\pm$2.2)          & 37.8($\pm$1.4) & \multirow{4}{*}{\makecell[c]{50.29 / 109.48M}} \\ 
Layer (4-6)   & \textit{41.9($\pm$1.4)}       & 43.1($\pm$1.8)           & \textbf{36.6($\pm$1.9)}          & \textbf{41.2($\pm$1.6)} &  \\ 
Layer (7-9)   & 38.1($\pm$2.7)       & \textit{50.5($\pm$8.2)}           & 26.9($\pm$1.6)          & 35.8($\pm$2.3) &  \\ 
Layer (10-12) & 35.5($\pm$5.3)       & 28.9($\pm$11.4)           & 24($\pm$7.5)            & 29.3($\pm$3.4) &  \\ \hline
IM-BERT      & 39.2($\pm$4.1)       & \textbf{58.8($\pm$13.5)}  & \textit{33.8($\pm$6.2)} & \textit{39.2($\pm$0.8)} & \makecell[c]{134.1 /109.48M} \\ \hline 
\bottomrule[0.7pt]
\end{tabular}
}
}
\caption{\textbf{Results on SST-2 in AdvGLUE} This table evaluates the efficacy of IM-connection against various adversarial attacks, measuring accuracy at the word, sentence, and human-crafted example levels. The "All" column provides an overall performance metric. The bold highlights the best performance. The computational efficiency of each model is indicated by FLOPs and parameters. Each result is averaged over three runs.
}
\label{table3}
\end{table*}
\subsection{Robustness in Low Resource Scenarios}
PLMs are vulnerable to adversarial attacks due to scarce data in downstream tasks. With fewer resources, the model becomes more susceptible. IM-BERT demonstrates its robustness in this challenging scenario.\\
\textbf{Setup} To evaluate the robustness of models in harsh situations like a low-resource training scenario, we sample 1000 and 500 instances from clean GLUE as train and validation sets and evaluate the models with the validation set of AdvGLUE. For a fair comparison, we conduct 3 runs with the same random seed in each run and report the average result. Other settings are the same as the main experiment.\\
\textbf{Result} The limited data of downstream tasks and the complexity of PLMs result in overfitting and vulnerability against adversarial attacks. The smaller the training dataset, the greater the vulnerability to overfitting and advertising attacks. Table 2 verifies that IM-connection robustifies the network in low-resource scenarios. With a sample of 1000 instances, the accuracies of IM-BERTs are consistently higher than those of the baseline for all tasks except for RTE. In the experiment with 500 samples, our models also maintain outperformance except for MNLI-mm and the RTE task. The gap between IM-BERT and baseline achieves higher accuracy up to 11.9\%p and 15.8\%p, respectively. Overall, IM-BERT with $T=10$ and $T=15$ performs better, 5.8\%p and 5.9\%p over the baseline, respectively. The gaps are more significant within 500 instances than in 1000, implying that our approach effectively robustifies BERT even in low-resource situations.
\subsection{Ablation Study}
To illustrate the effectiveness of our method for fine-tuning, we conduct additional experiments. These results indicate the effective placement of IM-connection and provide solutions to the inherent time-cost issues of implicit methods.

\noindent\textbf{How does IM-connection make BERT robust?}
As analyzed in the previous section, the implicit method effectively prevents the divergence of hidden states when perturbations are inserted into the initial value. To investigate how the IM-connection enhances the robustness of BERT against attacks at various levels, we construct 4 models with IM-connection applied at different positions within layers. We divide the 12 layers of BERT into 4 groups, fine-tune the models with SST-2, and test them with SST-2 in AdvGLUE. In this case, we train the models on the entire dataset and evaluate their performance, as shown in Table 1. To ensure fairness, we run 3 times and show the result as an average.

\noindent\textbf{Type of Attack Level} In Table 3, we analyze the effects of IM-connection and IM-BERT at various attack levels. Both methods enhance robustness by approximating the accurate hidden states of each token. They generally perform better in word-level perturbations and sentence-level adversarial attacks than in human-crafted examples, as they approximate hidden states for each token during prediction. This property leads to greater resistance to word-level adversarial attacks when the IM-connection is located in lower layers. Additionally, because PLMs learn semantics as they pass through layers, applying IM-connection to the middle layers helps mitigate perturbations, thereby preventing sentence-level attacks. Conversely, when the IM-connection is located in the middle layer, BERT becomes more robust against sentence-level adversarial attacks due to its improved ability to learn semantic representations. This is relevant to where the IM-connection conveys well-converged values on hidden states.

\noindent\textbf{Location of IM-connection} Table 3 reveals that placing the IM-connection in higher layers leads to poorer performance compared to lower layers. Specifically, when the IM-connection is positioned in higher layer groups like Layer (10-12), it fails to defend effectively against adversarial attacks. This suggests that BERT's earlier layers struggle to converge well on hidden states without the IM-connection, transmitting divergent values. In contrast, lower layers with the IM-connection can better approximate and deliver stable hidden states when the input is perturbed, successfully mitigating the attack. In other words, a well-approximated hidden state can be delivered via the IM-connection located in the lower layers, enhancing robustness against adversarial attacks.

\noindent\textbf{Strategic Application Based on Efficiency} Table 2 in the previous section shows the trade-off between time latency and robustness due to the inherent characteristics of the IM-connection. The experiment reveals that applying the IM-connection selectively to Layers (4-6) offers better performance than applying it to all layers, as seen in IM-BERT. This strategic approach, informed by an understanding of the PLM's structure, results in 2.25 times fewer FLOPs than IM-BERT, effectively addressing the time cost issue associated with the implicit method. This strategy opens up the possibility of an efficient application method that aims to maintain the number of parameters of the IM-connection while enhancing robustness. Additional analysis on BERT and RoBERTa regarding the trade-off and strategy is provided in Appendix A.

\subsection{Evaluating the Inherent Robustness}
Adversarial training methods require hyperparameter tuning based on the data and are challenging to adapt to various perturbations since they train on specific perturbations ~\cite{tramer2019adversarial}. This limitation underscores the necessity of improving a model’s inherent robustness. Our additional experiments demonstrate that IM-connection enhances inherent robustness through simple standard training, making it robust against various perturbations.

Table 4 shows the average performance of RoBERTa-based models across all tasks in the AdvGLUE dataset, broken down by each attack level. The results of the RoBERTa-based experiments confirm that IM-connection provides consistent robustness against various levels of perturbation. This demonstrates that unlike adversarial training—which is tailored to specific types of perturbations—IM-connection enhances the model's inherent robustness, enabling it to effectively handle a wide range of adversarial attacks.

\begin{table}[h!]
\setlength{\tabcolsep}{5pt}
\renewcommand{\arraystretch}{1.3}
\centering
\resizebox{\columnwidth}{!}{%
\LARGE
\begin{tabular}{c|ccc|c}
\toprule[1.5pt]
\hline
 & \multicolumn{3}{c|}{\begin{tabular}[c]{@{}l@{}}Adversarial Training\end{tabular}} & \multicolumn{1}{l}{\begin{tabular}[c]{@{}l@{}}Standard Training\end{tabular}} \\ \hline
\multicolumn{1}{c|}{Attack-Level} & \multicolumn{1}{c}{SMART} & \multicolumn{1}{c}{FreeLB} & \multicolumn{1}{c|}{InfoBERT} & \multicolumn{1}{c}{IM-RoBERTa} \\ \hline
\multicolumn{1}{c|}{Word-Level} & \textbf{62.01} & 52.29 & 55.42 & \textit{56.3} \\
\multicolumn{1}{c|}{Sentence-Level} & \textbf{47.72} & 44.24 & 38.03 & \textit{46.44} \\
\multicolumn{1}{c|}{Human Crafted} & 28.69 & \textbf{46.12} & 33.74 & \textit{41.89} \\ \hline
\multicolumn{1}{c|}{Avg.} & 46.14 & \textit{47.55} & 42.4 & \textbf{48.21} \\
\hline \bottomrule[1.5pt]
\end{tabular}%
}
\caption{\textbf{Performance comparison of RoBERTa-based models on the AdvGLUE dataset across different attack levels} The table shows the average accuracy of models trained with adversarial training methods (SMART, FreeLB, InfoBERT) versus standard training with IM-connection (IM-RoBERTa). The results are broken down by attack level (Word-Level, Sentence-Level, Human Crafted) and the overall average. The best performance for each category is highlighted in bold, and the second-best is in italic}
\end{table}

\section{Conclusion}
In this paper, we propose an approach to enhance PLMs with a more robust architecture by interpreting them as an ODE solver. Through our analysis of  the numerical stability, we find that the implicit method exhibits absolute stability against initial value perturbations regardless of the step size $\gamma$. Leveraging this property, we introduce IM-connection using the gradient descent method to approximate the solution of the implicit method. IM-connection effectively captures well-approximated hidden states without increasing the number of parameters. 
Our experimental results demonstrate the effectiveness of our approach on BERT. By simply modifying BERT to IM-BERT with incorporating IM-connection, we fine-tune IM-BERT in standard training. IM-BERT becomes more resistant to various adversarial attacks, outperforming both baseline and other adversarial training methods in the AdvGLUE dataset. Furthermore, in low-resource scenarios where the model is prone to overfitting on limited data, our method showcases its effectiveness by achieving higher accuracy compared to BERT.

\section*{Limitation}
Although we improve the model's intrinsic robustness, challenges remain, particularly the high time cost associated with the implicit method. Our experiments suggest ways to reduce this time cost without compromising robustness. One approach is to apply the IM-connection selectively across multiple layers, optimizing the balance between time-cost and performance for different PLM architectures. While this study focuses on BERT-based models, future work will extend to decoder-only and encoder-decoder architectures. Additionally, we will explore alternative numerical methods, such as gradient descent, to address time-cost issues more efficiently and improve robustness across various PLMs.

\section*{Ethics Statement}
Our research upholds ethical standards by utilizing publicly available datasets for model fine-tuning, ensuring transparency and avoiding sensitive data. We build upon openly released pre-trained models, enhancing accessibility and transparency. By creating a model that effectively defends against adversarial attacks, we contribute to a safer and more resilient AI system. Our objective is to democratize advanced AI techniques, including our proposed approach, making them accessible to researchers across different resource levels and fostering inclusivity in the field.

\section*{Acknowledgment}
This research was supported by the Basic Science Research Program (NRF-2022R1C1C1008534) and the National R\&D Program (2021M3H2A1038042) through the National Research Foundation of Korea (NRF), funded by the Ministry of Education and the Ministry of Science and ICT, respectively. It was also supported by the Institute for Information \& Communications Technology Planning \& Evaluation (IITP) through the Korea government (MSIT) under Grant No. 2021-0-01341 (Artificial Intelligence Graduate School Program, Chung-Ang University).


\bibliography{custom}
\bibstyle{acl_natbib}
\section{Appendix}

\appendix
\renewcommand{\thesection}{\Alph{section}}
\setcounter{section}{0}

\section{Robustness and Time Latency with Iteration \textit{T}}
As a numerical approach, we utilize the gradient descent method to solve hidden states, causing iterative computations of \textit{T} for each layer.\\ To analyze our assumption, the trade off between iteration \textit{T} and performance, we conduct additional experiment.

\subsection{Robustness with diverse Iteration \textit{T}}
We conduct experiments utilizing TextFooler~\cite{23} and IMDB~\cite{42}. We train IM-BERT on the IMDB dataset and evaluate against adversarial attack generated by TextFooler, varying \textit{T} from 1 to 15. In Table 4, as the iteration \textit{T} increases, the result indicates that IM-connection enhances the network’s stability against adversarial attacks, with accuracy peaking at $T=10$. When $T=10$, the result exhibits better accuracy with fewer FLOPs than $T=15$. This observation suggests that we can identify optimal iteration \textit{T} that maximizes performance for the specific dataset.

\begin{table}[h]
\setlength{\tabcolsep}{10pt}
\renewcommand{\arraystretch}{1.5}

\centering
\resizebox{\columnwidth}{!}{%
\small
\begin{tabular}{c|cccc}
\toprule[1.0pt]
\hline
TextFooler & \textit{T}=1 & \textit{T}=5 & \textit{T}=10  & \textit{T}=15 \\ \hline
Accuracy & 0.2228 & 0.2316 & \textbf{0.2747} & 0.2633 \\ 
FLOPs&  44.7& 134.1 &245.85 &357.59\\ \hline
\bottomrule[1.0pt]
\end{tabular}%
}
\caption{\textbf{Results on IMDB attacked by TextFooler} we report Accuracy and FLOPs to evaluate the performance according to \textit{T} in IM-connection}
\label{table4}
\end{table}

\subsection{Strategic IM-connection Application to mitigate time-latency issue}
\begin{table}[h!]
\setlength{\tabcolsep}{10pt}
\renewcommand{\arraystretch}{1.5}
\centering
\resizebox{\columnwidth}{!}{%
\small
\begin{tabular}{l|c|c|c}
\toprule[0.7pt]
\hline
 & Accuracy & FLOPs & Params \\ \hline
BERT & 29.1 & 22.35 & \multirow{7}{*}{109.48M} \\ \cline{1-3}
SMART$_\textit{BERT}$ & 20.6 & 67.09 &  \\ \cline{1-3}
IM-BERT(1-3) & 37.8 & \multirow{4}{*}{50.29} &  \\
IM-BERT(4-6) & \textbf{41.2} &  &  \\
IM-BERT(7-9) & 35.8 &  &  \\
IM-BERT(10-12) & 29.3 &  &  \\ \cline{1-3}
IM-BERT(All) & \textit{39.2} & 134.1 &  \\  \toprule[0.7pt] \bottomrule[0.8pt]
SMART$_\textit{RoBERTa}$ & 50.9 & 236.88 & \multirow{4}{*}{355.36M} \\ \cline{1-3}
IM-RoBERTa(4-7) & {\ul 53.4} & \multirow{2}{*}{144.76} &  \\
IM-RoBERTa(11-14) & \textbf{54.7} &  &  \\ \cline{1-3}
IM-RoBERTa(All) & 52 & 473.75 & \\ \hline
\bottomrule[0.8pt]
\end{tabular}%
}
\caption{\textbf{Result on SST-2 in AdvGLUE with BERT and RoBERTa} This table evaluates the efficacy of IM-connection against adversarial attack on SST-2 in AdvGLUE. The bold highlights the best performance. The computational efficiency of each model is indicated by FLOPs and Params.}
\label{table5}
\end{table}
While the gradient descent approach is essential for solving the implicit method, it introduces time latency, evident from Table 4's linear increase in latency according to Algorithm 1. To mitigate this, Table 3 suggests applying IM-connection to specific layers. Incorporating IM-connection in early or middle layers alleviates time latency and ensures adversarial robustness. The following Table 6 reinforces the benefits of this strategic application.

Table 6 illustrates the effectiveness of strategically applying the IM-connection into specific layers in both BERT and RoBERTa models. The results demonstrate that selectively incorporating IM-connection in certain layers,  particularly in the middle layers (e.g., Layers 4-6 for BERT and Layers 11-14 for Roberta), improves accuracy while managing computational efficiency (FLOPs) and the number of parameters.

When IM-connection is applied to all layers, IM-RoBERTa demands roughly 2.00x the FLOPs of SMART, indicating a significant increase in computational cost without a corresponding boost in adversarial robustness. However, selectively applying IM-connection to middle layers, like Layers 4-6 in BERT and 11-14 in RoBERTa, uses resources more efficiently. This selective approach requires 1.64x fewer FLOPs than the all-layer IM-RoBERTa and surpasses SMART in both robustness and accuracy. The strategic layer-specific application of IM-connection effectively balances computational efficiency with enhanced adversarial resistance. This method highlights the benefits of focusing on the most impactful layers to optimize performance.

\end{document}